\documentclass[conference]{IEEEtran}
\IEEEoverridecommandlockouts

\usepackage{cite}
\usepackage{amsmath,amssymb,amsfonts}
\usepackage{algorithmic}
\usepackage{graphicx}
\usepackage{textcomp}
\usepackage{xcolor}
\usepackage{wrapfig}
\usepackage{algorithm}
\usepackage{subfigure}
\usepackage{adjustbox}
\usepackage{MnSymbol}
\usepackage{multirow}
\usepackage{float}
\usepackage{everyshi}
\usepackage{booktabs}
\usepackage{enumitem}
\def\BibTeX{{\rm B\kern-.05em{\sc i\kern-.025em b}\kern-.08em
    T\kern-.1667em\lower.7ex\hbox{E}\kern-.125emX}}
\begin{document}

\title{DICS: Find Domain-Invariant and Class-Specific Features for Out-of-Distribution Generalization\\
\thanks{This work was supported by the National Natural Science Foundation of China (62293554, 62206249, U2336212), Natural Science Foundation of Zhejiang Province, China (LZ24F020002), and Young Elite Scientists Sponsorship Program by CAST (2023QNRC001).}
}

\author{\IEEEauthorblockN{1\textsuperscript{st} Qiaowei Miao}
\IEEEauthorblockA{\textit{Zhejiang University} \\
Hangzhou, China \\
qiaoweimiao@zju.edu.cn}
\and
\IEEEauthorblockN{2\textsuperscript{nd} Yawei Luo \IEEEauthorrefmark{2}} 
\IEEEauthorblockA{\textit{Zhejiang University} \\
Hangzhou, China \\
yaweiluo@zju.edu.cn}
\and
\IEEEauthorblockN{3\textsuperscript{rd} Yi Yang}
\IEEEauthorblockA{\textit{Zhejiang University} \\
Hangzhou, China \\
yangyics@zju.edu.cn}

}

\maketitle

%
\begin{abstract}
While deep neural networks have made remarkable progress in various vision tasks, their performance typically deteriorates when tested in out-of-distribution (OOD) scenarios. Many OOD methods focus on extracting domain-invariant features but neglect whether these features are unique to each class. Even if some features are domain-invariant, they cannot serve as key classification criteria if shared across different classes. In OOD tasks, both domain-related and class-shared features act as confounders that hinder generalization.
In this paper, we propose a \textbf{DICS} model to extract \textbf{D}omain-\textbf{I}nvariant and \textbf{C}lass-\textbf{S}pecific features, including  Domain Invariance Testing (DIT) and Class Specificity Testing (CST), which mitigate the effects of spurious correlations introduced by confounders. DIT learns domain-related features of each source domain and removes them from inputs to isolate domain-invariant class-related features. DIT ensures domain invariance by aligning same-class features across different domains. Then, CST calculates soft labels for those features by comparing them with features learned in previous steps. We optimize the cross-entropy between the soft labels and their true labels, which enhances same-class similarity and different-class distinctiveness, thereby reinforcing class specificity. Extensive experiments on widely-used benchmarks demonstrate the effectiveness of our proposed algorithm. Additional visualizations further demonstrate that DICS effectively identifies the key features of each class in target domains.

\end{abstract}
\begin{IEEEkeywords}
Out-of-distribution Generalization,  Invariant Representation Learning, Causal Inference.
\end{IEEEkeywords}
\section{Introduction}
\label{sec:intro}
Deep learning methods have demonstrated exceptional progress in various fields over the past few years. However, the deep learning model usually suffers from performance shutdown when the test target domain has a different distribution from the training data~\cite{rame2021fishr,krueger2021out,miao2024domaindiff}.  This issue severely hinders the applicability of deep models in practical settings.

\begin{figure}[t]
    \centering
    \includegraphics[width=1\linewidth]{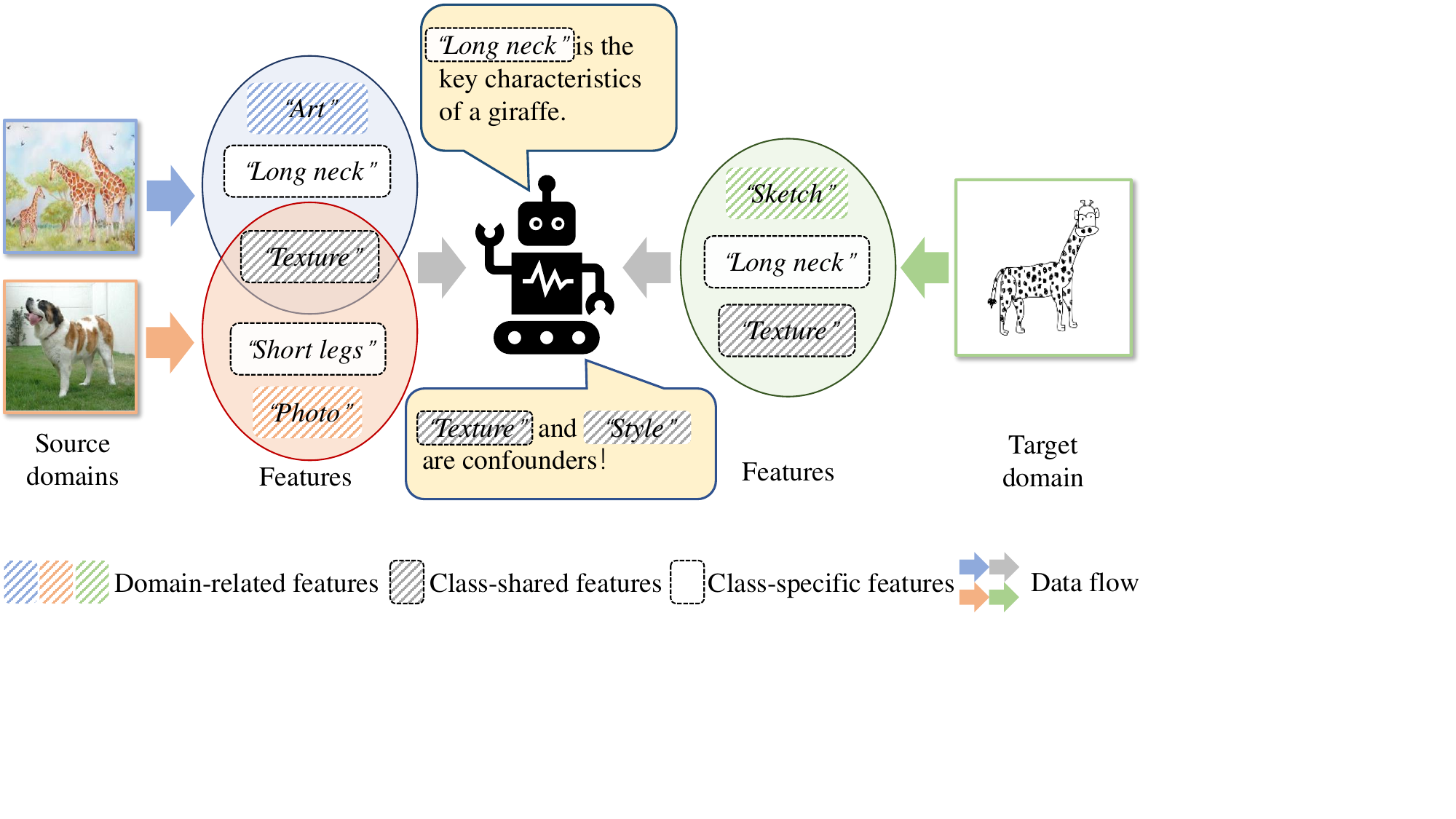}
     \vspace{-0.75cm}
    \caption{Domain-related and class-shared features are confounders that undermine models' out-of-distribution generalization. A model might mistakenly classify an image from the target domain as a dog due to similar textures or because its sketch style more closely resembles art than photographic styles. Our goal is to eliminate the influence of these confounders and identify domain-invariant, class-specific features that genuinely define each class, such as the giraffe's long neck.}
    \label{fig:intro}
     \vspace{-0.5cm}
\end{figure}

To improve generalization performance, numerous invariant representation learning methods~\cite{rame2021fishr,arjovsky2019invariant,li2018deep,chevalley2022invariant}  have been proposed.
Those methods endeavor to extract domain-invariant features as the basis for prediction. Nevertheless, such features may not capture the unique information of each class well enough for classification. As shown in Fig.~\ref{fig:intro}, both "texture" and "long neck" are domain-invariant and class-related features of giraffes. However, only the "long neck" is key for classifying a giraffe. The "texture", shared by the dog and giraffe, acts as a confounder that can mislead the model into predicting the input as a dog because dogs also have similar textures.
If the model relies on class-shared features, like "texture",  to classify data during the training phase and then attempts to use the same features for prediction in the testing data, this reliance on features that lack true invariance can harm the model's generalization ability.

\begin{wrapfigure}{r}{0.35\linewidth}
 \vspace{-10pt}
  \begin{center}
    \includegraphics[width=\linewidth]{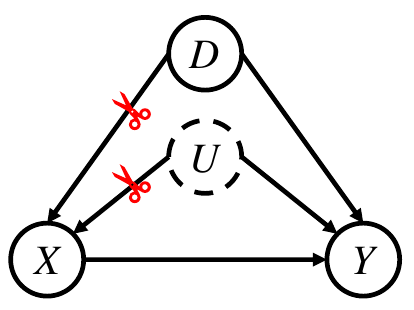}
  \end{center}
  \vspace{-10pt}
  \caption{SCM of OOD task.}
   \vspace{-10pt}
  \label{fig:scm}
\end{wrapfigure}

In a causal view~\cite{bunge2017causality}, we introduce a structural causal model (SCM)~\cite{scholkopf2022causality} to formalize the OOD problem~\cite{wu2023learning,miao2022domain,mahajan2021domain,chen2021style }. As shown in Fig.~\ref{fig:scm}, $X \rightarrow Y$ means the unique semantic part of input $X$  causes their labels $Y$, which is the genuine causal mechanism and is invariant~\cite{arjovsky2019invariant,scholkopf2021toward,scholkopf2022statistical,bunge2017causality, chen2021style}. 
$D$ is domain-related features (e.g. domain style) and $U$ is domain-invariant but class-shared features  (e.g. "texture"). 
The presence of $D$ and $U$ introduces two additional spurious associations (e.g., $X \leftarrow D \rightarrow Y$ and $X \leftarrow U \rightarrow Y$ ) between image $X$ and label $Y$, which change separately in response to variations across different domains and different classes, hurting the OOD performance of models. Based on our SCM,  previous invariant representation learning methods~\cite{chevalley2022invariant,li2021invariant,arjovsky2019invariant,liu2022investigating} remove the effects of domain-related confounder $D$ is not enough, the spurious associations caused by class-shared confounder $U$ also need to be cut off. Thus, We need to find domain-invariant and class-specific features for predictions, which have true causal relationships with labels, corresponding to the causal path $X \rightarrow Y$.

In this paper, we propose a model named \textbf{DICS} to extract \textbf{D}omain-\textbf{I}nvariant and \textbf{C}lass-\textbf{S}pecific features as the key classification to enhance the OOD performance. DICS includes Domain Invariance Testing (DIT) and Class Specificity Testing (CST). To mitigate domain effects, DIT learns and removes domain-specific features from each source domain to extract domain-invariant and class-related features. Additionally, DIT computes the similarity of extracted features of the same class across different domains to assess and enhance domain invariance by maximizing these similarities.
To ensure class-specific features, CST compares the input with historical knowledge to discern class differences. We use an invariant memory queue to store learned features and their true labels. By computing the similarity matrix between current class-related features and those in the queue, we derive a soft label through a weighted summation of true labels. Optimizing the cross-entropy loss between the soft label and the true label enhances intra-class similarity while minimizing inter-class similarity, ensuring that domain-invariant features remain class-specific.
We evaluate DICS on multiple datasets, including PACS, OfficeHome, TerraIncognita, and DomainNet. DICS performs competitively with state-of-the-art methods in terms of accuracy. We summarize our contributions as follows:

\begin{itemize}[itemsep=0pt, topsep=0pt]
    \item 
    We rethink invariant representations learning in the OOD task and highlight that domain-invariant and class-shared features act as confounders that hurt models' OOD generalization performance.
    \item We propose the DICS model, which integrates domain invariance testing to ensure the consistency of same-class features after removing domain-related elements. Additionally, DICS utilizes class specificity testing to mitigate excessive similarity between non-same-class features, thereby preserving class-specific distinctions.
    \item Extensive experiments on PACS, OfficeHome, TerraIncognita, and DomainNet demonstrate the superior performance of DICS in OOD task.
\end{itemize}

\section{Related Works}

Out-of-Distribution (OOD) generalization aims to train a model using multiple source domains that can generalize well to unseen target domains~\cite{meinshausen2015maximin,rothenhausler2021anchor, sun2020test, zhang2021deep}. OOD methods can be classified into three categories: data manipulation, learning strategies, and representation learning. Data manipulation~\cite{volpi2018generalizing, shankar2018generalizing, Carlucci2019DomainGB, Wang2020LearningFE, DBLP:conf/aaai/ZhouYHX20, zhou2020learning, zhou2021domain} techniques involve modifying or generating input data to facilitate generalization. Learning strategies, exemplified by ensemble learning and meta-learning, aim to enhance generalization using commonly applicable learning techniques. Meta-learning~\cite{balaji2018metareg, li2018learning, dou2019domain, Li2019EpisodicTF, li2019feature} involves partitioning training data and simulating domain variations. Representation learning is a prominent research area in OOD, focusing on minimizing distribution discrepancies within the training domain using methods such as adversarial learning~\cite{DBLP:conf/aaai/ZhouYHX20, zhou2020learning} and representation alignment~\cite{Zhao2020DomainGV, Matsuura2020DomainGU, li2018deep, DBLP:conf/aaai/LiGTLT18}, to learn domain-invariant features. However, the invariant representations learned from the source domain do not all necessarily conform to causal relationships~\cite{mahajan2021domain}. 

Invariant representation learning tends to focus on removing domain-related features while neglecting class-shared features that are invariant across source domains but irrelevant to the target domain~\cite{nguyen2023causal,chen2021style }. Our work not only introduces a domain invariance test to assess the domain invariance of extracted features but also proposes a class specificity test to ensure that the features capture key information relevant to each category. From a causal perspective, our method builds upon traditional statistical learning by eliminating both domain-related and class-related spurious associations, thereby aiding the model in identifying crucial classification cues and enhancing its performance in OOD tasks.

\section{Method} 

\begin{figure}
    \centering
    \includegraphics[width=\linewidth]{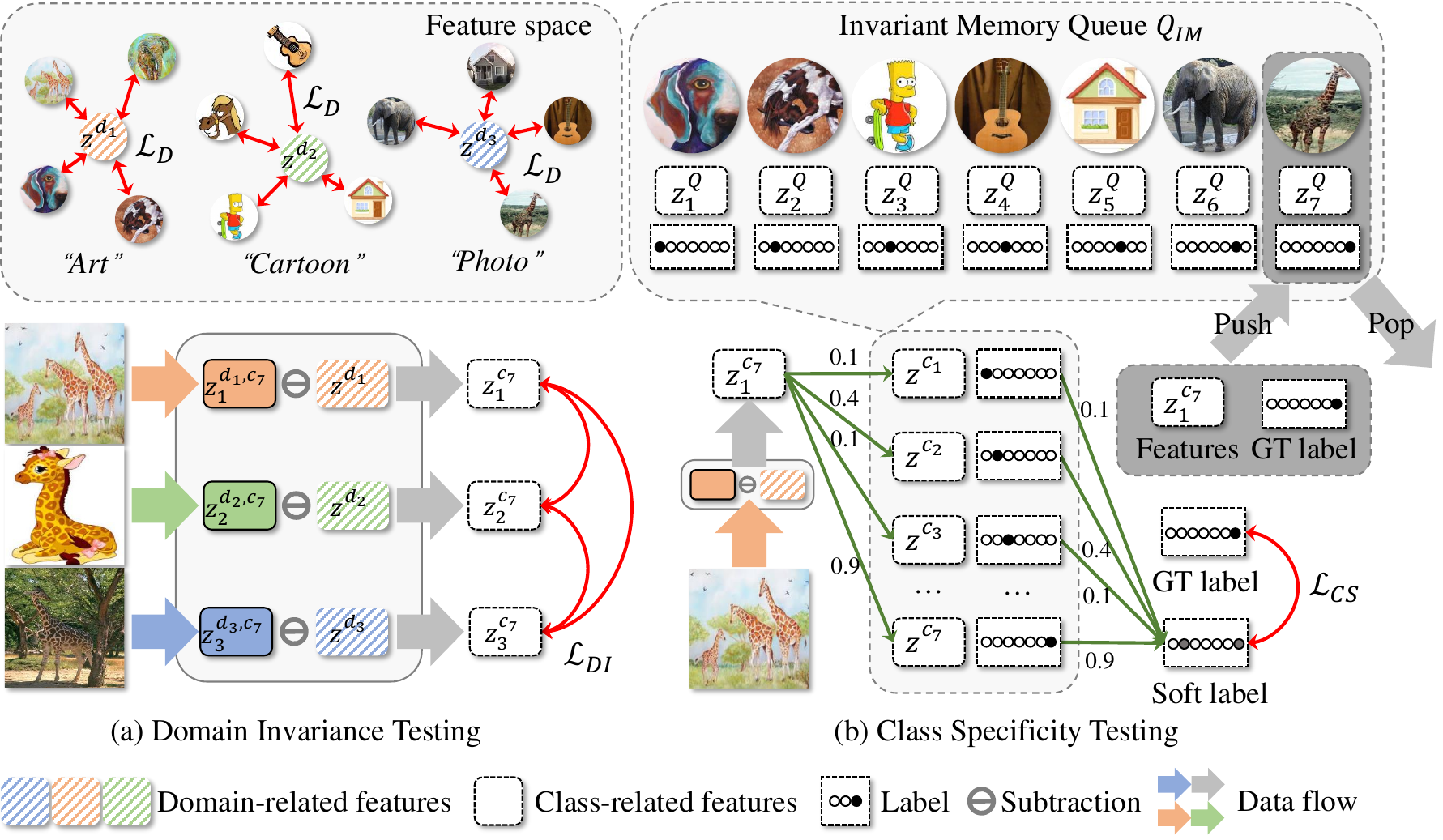}
     \vspace{-0.75cm}
    \caption{DICS model includes domain invariance testing and class specificity testing. The former maximizes the similarity of class features across different source domains to ensure domain invariance. The latter forces the current input's class features to be closer to those of the same class and farther from those of other classes to maintain class specificity.}
    \label{fig:all}
     \vspace{-0.5cm}
\end{figure}

\subsection{Preliminary}
OOD task aims to train a generalizable model from $D$ source domains and test its performance on an unseen target domain.  We build upon the foundation of   Empirical Risk Minimization (ERM)~\cite{vapnik1999nature}, which has a pipeline that image encoder $E$ extract features $z$ from an input $x$, and sends $z$ into a classifier $F$ to predict its label $y'$. By minimizing cross-entropy loss $\mathcal{L}_{C}$ as optimization target:
\begin{equation}
    \mathcal{L}_{C} = -\frac{1}{N}\sum_{i=1}^{N}y_{i} \textrm{log}( F(E(x_{i}))),
    \label{loss:C}
\end{equation}
where $N$ is the batch size. 
Based on ERM, DICS proposes Domain Invariance Testing(DIT) and Class Specificity  Testing (CST) to make the extracted features $z$  satisfy domain invariance and class specificity.

\subsection{Domain Invariance Testing}
The samples from the same domain share some domain-related features, thus DIT learns such domain-related features and removes them from extracted features to make them domain-invariant. Especially, we randomly sample $N_{d}$ images from each source domain to construct a batch. To learn domain-related features for domain $d$, we construct learnable vectors $z^{d}$ as domain-related features and maximize the similarity between $z^{d}$ and the samples' features $z^{d,c}_{i}$ from domain $d$ in the current batch. The similarity $\mathrm{sim}(z^{d},z^{d,c}_{i})$ is as follows:
\begin{equation}
    \mathrm{sim}(z^{d},z^{d,c}_{i})=\frac{\textrm{norm}(z^{d}) \cdot \textrm{norm}(z^{d,c}_{i})^{T}}{\tau \cdot \sqrt{\textrm{d}}},
\end{equation}
where $\tau$ is a temperature hyperparameter and $\textrm{d}$ is the dimension of the features. Then,
we optimize $\mathcal{L}_{D}$ for updating $z^{d}$:
\begin{equation}
   \mathcal{L}_{D} =  -\frac{1}{D }\sum_{d=1}^{D}  \frac{1}{N_{d}}  \sum_{i=1}^{N_{d}}  \textrm{log}(\textrm{sim}(z^{d}, z^{d,c}_{i})),
\end{equation}
where $N_{d}$ is the number of samples from domain $d$.
For a sample $x_{i}$, its extracted features $z_{i}^{d,c}$ are mixtures of domain-related features and class-related features. Therefore, we subtract $z^{d}$ from $z_{i}^{d,c}$ to isolate class-related features $z_{i}^{c}$. To ensure $z_{i}^{c}$ are domain-invariant, we maximize the similarity between features with the same class but from different domains by optimizing domain-invariant loss $\mathcal{L}_{DI}$:
\begin{equation}
    \mathcal{L}_{DI} =  -\frac{1}{C}\sum^{C}_{c=1}\frac{2}{N_{c}(N_{c}-1) }  \sum_{i=1}^{N_{c}}\sum_{j=i+1}^{N_{c}}  \textrm{log}(\textrm{sim}(z^{c}_{i}, z^{c}_{j})),
\end{equation}
where $N_{c}$ is the number of samples belonging to class $c$ in one batch and $C$ is the number of classes.

\subsection{Class Specificity  Testing}
After getting domain-invariant and class-related features $z^{c}_{i}$, we need to test whether they are unique enough to represent its class. We propose Class Specificity  Testing (CST), which tests whether the features $z^{c}_{i}$ are similar to other same-class features and different from non-same-class features. However, the sample size in a single batch is too small to objectively assess the differences between features, while using all training data incurs excessive time costs. Therefore, we build an invariant memory queue $Q_{IM}$ to store an appropriate number of features trained from previous steps and provide features to compare. As shown in Fig.~\ref{fig:all} (b), based on the similarity between current features $z_{i}^{c}$ and features $z^{Q}_{j}$ stored in $Q_{IM}$, we can obtain the corresponding soft label $\hat{y}_{i}$ by performing a weighted sum of the one-hot labels from multiple classes. We optimize class-specific loss $\mathcal{L}_{CS}$:
\begin{equation}
    \mathcal{L}_{CS} = -\frac{1}{N}\sum_{i=1}^{N}y_{i} \textrm{log}(\sum_{j=1}^{N_{Q}}y_{j}\textrm{sim}(z_{i}^{c}, z^{Q}_{j})). 
    \label{loss:CS}
\end{equation}
$\mathcal{L}_{CS}$ penalizes instances where different classes have similar features while encouraging similarity within the same class to ensure domain-invariant features $z^{c}_{i}$ are also class-specific.

\subsection{Optimization}
In each step, we optimize $L_{D}$ to learn domain-related features $z^{d}$ for each source domain $d$ first. Then, we froze $z^{d}$ and train our model. The overall optimization objective $\mathcal{L}_{DICS}$ of DICS can be summarized as follows:
\begin{equation}
    \mathcal{L}_{DICS} =  \mathcal{L}_{C} + \alpha \mathcal{L}_{DI} + \beta \mathcal{L}_{CS},
    \label{eq:overall}
\end{equation}
where $\alpha$ and $\beta$ are hyperparameters discussed in the next experiments.
We optimize $\mathcal{L}_{DICS}$ to force our model to find domain-invariant and class-specific features for prediction. after each training step, we pop the oldest features in $Q_{IM}$ and push current features $z^{c}_{i}$ and corresponding label into $Q_{IM}$. However, training updates could still introduce interference with extracted and will be poped features $z^{c}_{i}$. We  slowly update $E$  to maintain consistency:
\begin{equation}
    \theta_{E} \leftarrow \lambda \theta_{E} + (1-\lambda)\theta_{E'},
\end{equation}
where $\theta_{E}$ is the last step weights and $\theta_{E'}$ is updated weights of encoder $E$. $ \lambda$ is the fusion coefficient.

\section{Experiments}

\begin{table}[t]
\caption{Comparison with DICS and other methods.}
 \vspace{-0.8cm}
\setlength{\tabcolsep}{0.5mm}
\begin{center}
\adjustbox{max width=\linewidth}{%
\begin{tabular}{lcccccccc}
\toprule
\multicolumn{6}{l}{ Model selection: "Training-domain" validation set} \\
\midrule
\textbf{Algorithm}        & \textbf{PACS}             & \textbf{OfficeHome}       & \textbf{TerraIncognita}   & \textbf{DomainNet}       & \textbf{Avg.}              \\
\midrule
ERM~\cite{vapnik1999nature}                        & 85.5            & 66.5            & 46.1            & 40.9             & 59.8                      \\
IRM~\cite{arjovsky2019invariant}                         & 83.5            & 64.3            & 47.6            & 33.9             & 57.3                      \\
GroupDRO~\cite{sagawa2019distributionally}             & 84.4            & 66.0            & 43.2            & 33.3             & 56.7                      \\
Mixup~\cite{yan2020improve}                     & 84.6            & 68.1            & 47.9            & 39.2             & 60.0                      \\
MLDG~\cite{li2018learning}                      & 84.9            & 66.8            & 47.7            & 41.2             & 60.2                      \\
CORAL~\cite{sun2016deep}                    & 86.2            & 68.7            & 47.6            & 41.5             & 61.0                     \\
MMD~\cite{li2018domain}                       & 84.6            & 66.3            & 42.2            & 23.4             & 54.1                    \\
DANN~\cite{ganin2016domain}                      & 83.6            & 65.9            & 46.7            & 38.3             & 58.6                      \\
CDANN~\cite{li2018deep}                    & 82.6            & 65.8            & 45.8            & 38.3             & 58.1                      \\
MTL ~\cite{blanchard2017domain}                        & 84.6            & 66.4            & 45.6            & 40.6             & 59.3                      \\
SagNet~\cite{nam2021reducing}                     & 86.3            & 68.1            & \underline{48.6}            & 40.3             & 60.8                     \\
ARM~\cite{zhang2020adaptive}                         & 85.1            & 64.8            & 45.5            & 35.5             & 57.7                     \\
VREx~\cite{krueger2021out}                        & 84.9            & 66.4            & 46.4            & 33.6             & 57.8                     \\
RSC~\cite{huang2020self}                           & 85.2            & 65.5            & 46.6            & 38.9             & 59.1                      \\
AND-mask~\cite{shahtalebi2021sand}               & 84.4            & 65.6            & 44.6            & 37.2            &  58.0                         \\
SAND-mask~\cite{shahtalebi2021sand}             & 84.6            & 65.6            & 42.9            & 32.1            &  56.3                   \\
Fishr~\cite{rame2021fishr}                         & 85.5            & 67.8            & 47.4            & 41.7            &  60.6                     \\
EQRM~\cite{eastwood2022probable}                        & 86.5            & 67.5            & 47.8            & 41.0            &  60.7                    \\
CausIRL-MMD~\cite{chevalley2022invariant}         & 84.0            & 65.7            & 46.3            & 40.3            &   59.1                   \\
CausIRL-CORAL~\cite{chevalley2022invariant}      & 85.8            & 68.6            & 47.3            & 41.9            &  60.9       \\
CB-CORAL~\cite{wang2022causal}                 & 86.7           & \underline{69.6}             & 47.0            & \underline{43.9}           &  \underline{61.8}            \\
ADRMX~\cite{demirel2023adrmx}                        & 85.3     & 68.3   & 47.4    & 43.1     & 61.0  \\  
RDM~\cite{nguyen2024domain}                        & \underline{87.2}     & 67.3   & 47.5    & 43.4     & 61.4  \\  
\midrule        
DICS                              & \textbf{88.4}   & \textbf{70.6}   & \textbf{50.4}   & \textbf{44.1}    & \textbf{63.4}      \\
\bottomrule
\end{tabular}}
\end{center}
\label{tab:all}
 \vspace{-0.75cm}
\end{table}

\hspace{1pt}\textbf{Implementation details.} 
We use ResNet50 as the backbone to get results with the settings following $\mathrm{DomainBed}$~\cite{gulrajani2020search}. The temperature hyper-parameter $\tau$ is 0.07, and the momentum ratio $\lambda$ is 0.999, the same as MoCo~\cite{he2020momentum}. We use RTX 3090 $\times$ 2 to support the derivation of results. Each RTX 3090 graphics card has 24 GB of memory. The version of PyTorch is 1.10.0. We train DICS  over 3 times in each dataset. 

\textbf{Datasets.}
PACS~\cite{li2017deeper} contains 9,991 images with 4 domains $\{art, cartoon, photo,sketch \}$ and 7 categories.
OfficeHome~\cite{venkateswara2017deep} contains 15,588 images with 4 domains $\{art, clipart, product,real\mbox{-}world \}$ and 65 categories.
The settings of TerraIncognita~\cite{venkateswara2017deep} remain the same as 
\cite{gulrajani2020search}. It contains 24,788 images with 4 domains $\{L100, L38, L43, L46 \}$ and 10 categories.
DomainNet~\cite{XingchaoPeng2018MomentMF} contains 569,010 images with 6 domains $\{clip, info, paint, quick, real, sketch \}$ and 345 classes.

\textbf{Results on DG datasets.} 
In Table~\ref{tab:all}, DICS simultaneously achieves the best performance on  PACS~\cite{li2017deeper} , OfficeHome~\cite{venkateswara2017deep}, TerraIncognita~\cite{venkateswara2017deep} and DomainNet~\cite{XingchaoPeng2018MomentMF}. 
We highlight the \textbf{best results} and the \underline{second best results}.
\textbf{PACS} has a noticeable stylistic difference and multiple similar classes.  DICS significantly outperforms the second-best model RDM, by a margin of 1.2 points. While RDM minimizes the variance of risk distributions across training domains to reduce the effects of domain shifts, it overlooks the class-shared features are also confounders that could confuse models to make mistakes.
\textbf{OfficeHome}, including multiple source domains, exhibit distinct stylistic differences, yet there are more classes compared to PACS. On OfficeHome, compared to  CB-CORAL, a model aims to reduce all spurious correlations in the dataset, DICS leads to improvements of 1.0 points respectively, which shows the advantages of DICS dealing with complex multi-domain, multi-class data.
\textbf{TerraIncognita} dataset exhibits significant background variations, which requires extract features to be domain-invariant.  While SagNet mitigates the impact of class-related confounders by swapping background information between images, CST helps DICS extract better features and achieve the best accuracy, outperforming SagNet by 1.8 points. 
\textbf{DomainNet}  comprises a larger number of classes than others, which asks the model to understand the differences between classes accurately.  While CB-CORAL gets an accuracy of 43.9, DICS can still outperform it.

\begin{table}[tbp]
    \caption{Ablation experiments of $\mathcal{L}_{DI}$ and $\mathcal{L}_{CS}$.}
    \vspace{-0.2cm}
    \label{tab:ablation}
    \centering
    \adjustbox{max width=\linewidth}{%
    \begin{tabular}{ccc|cc|c}
    \toprule
    \multicolumn{3}{c|}{\textbf{DICS}}     & \multirow{2}{*}{\textbf{PACS}}           & \multirow{2}{*}{\textbf{TerraIncognita}}                       & \multirow{2}{*}{\textbf{Avg.}}  \\
    \textbf{$\mathcal{L}_{C}$}   & \textbf{$\mathcal{L}_{DI}$ ($\alpha$)}  & \textbf{$\mathcal{L}_{CS} $($\beta$) }           & &  & \\
    
    \midrule
    1.0                       & 0.0  &  0.0                 & 85.5                                                    & 46.1                                                      &  65.8                      \\
    1.0           & 0.5  &  0.0                   & 86.9                                                 & 49.0                                           & 68.0  \\
    1.0           & 1.0  &  0.0                   & 87.6                                                 & 48.7                                & 68.2 \\
    1.0           & 0.0  &  0.5                   & 87.0                                                 & 48.3                                & 67.7 \\
    1.0           & 0.0  &  1.0                   & 87.8                                                 & 48.7                                & 68.3 \\
    1.0  &  0.5 &  1.0                 & 87.9                   &  49.4                                                                        & 68.1  \\
    1.0  & 1.0  & 0.5                  &   87.7             &    49.6                                                                         & 68.7 \\
    1.0 & 1.0  & 1.0 &     \textbf{88.4}    & \textbf{50.4}     & \textbf{69.4}      \\
    \bottomrule
    \end{tabular}}
     \vspace{-0.5cm}
\end{table}

\begin{table}[tbp]
    \caption{Comparative experiments of the length $N_{Q}$ of $Q_{IM}$. }
    \vspace{-0.2cm}
    \label{tab:len}
    \centering
    \adjustbox{max width=\linewidth}{%
    \begin{tabular}{lccc}
    \toprule
    \textbf{Algorithm}              & \textbf{PACS}                 & \textbf{TerraIncognita}                  & \textbf{Avg.}\\
    \midrule
    DICS\mbox{-}1$N$                    & 87.9                                              & 48.5                                                 & 68.2 \\
    DICS\mbox{-}4$N$                     & 88.4                                               & 50.4                                                & 69.4 \\
    DICS\mbox{-}8$N$                    & 88.0                                         & 50.0                                                       & 69.0\\ 
    DICS\mbox{-}16$N$                    & 85.9                                         & 50.2                                                     & 68.1\\ 
    \bottomrule
    \end{tabular}}
     \vspace{-0.5cm}
\end{table}

\textbf{Effects of $\mathcal{L}_{DI}$ and  $\mathcal{L}_{CS}$.}
We conduct ablation experiments on  PACS and TerraIncognita, the former tests domain invariance, while the latter requires the model to distinguish classes accurately.
As shown in Table~\ref{tab:ablation}, While the coefficient $\alpha$ of $\mathcal{L}_{DI}$ increases, the features extracted by DICS are more domain-invariant, thus the accuracy on PACS gradually get higher.  When only utilizing the CST, the accuracy of DICS increases monotonically on TerraIncognita as the coefficient $\beta$ of $\mathcal{L}_{CS}$ increases, 
When both $\alpha$ and $\beta$ are set to 1.0, DICS achieves the best results on both datasets.

\textbf{Sensitivity of the length of $Q_{IM}$.} The setting of $N_{Q}$ is a trade-off between training update interference and the breadth of class features. We update $Q_{IM}$ on a batch-by-batch basis, which means the length $ N_{Q}$ of $Q_{IM}$ is a multiple of the batch size $N$. We fix the seed and only change the $N_{Q}$. According to Table~\ref{tab:len}, $N_{Q}=4N$ strikes a good balance .

\begin{figure}[tbp]
\centering
\includegraphics[width=0.45\textwidth]{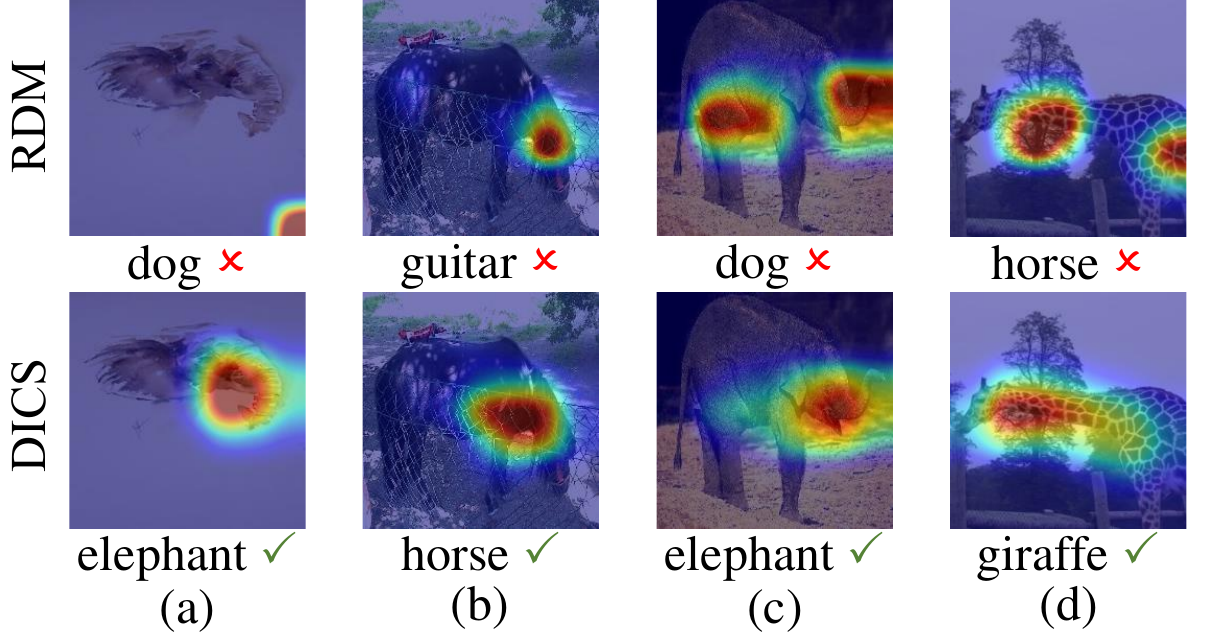}
 \vspace{-0.3cm}
\caption{Visual results of RDM and DICS on PACS. 
(a) The background serves as a domain-specific confounder.
(b) The fence in front of the horse is a feature shared with the "guitar" class, which confuses RDM.
(c) DICS focuses more on the elephant's trunk, the key to identifying elephants.
(d) DICS recognizes the giraffe's long neck as the basis for prediction.
}
\label{figs:gradcam}
 \vspace{-0.5cm}
\end{figure}

\textbf{Visualization results.} We use Grad-CAM~\cite{RamprasaathRSelvaraju2016GradCAMVE} to visualize attention differences between RDM and DICS on PACS.
In Fig.~\ref{figs:gradcam} (a), images from the "sketch" source domain have a significant amount of blank background, causing RDM to make incorrect predictions based on the background in the "art painting" target domain. However, DICS can accurately capture the region of the elephant's long trunk without being affected by the white background. 
In Fig.~\ref{figs:gradcam}(b), as horses and wire fences coincidentally appear together, RDM erroneously associates the wire fence with guitar strings, leading to the misclassification of the horse as a guitar. DICS extends its focus to the horse's mane and correctly predicts it as a horse.
In Fig.~\ref{figs:gradcam} (c) and Fig.~\ref{figs:gradcam} (d), RDM  focuses on multiple regions that do not serve as distinctive class markers, such as the elephant's legs and the giraffe's texture. DICS adjusts these regions to concentrate on class-specific features like the elephant's trunk and the giraffe's neck.

\vspace{-0.1cm}
\section{Conclusion}


In this paper, we propose a DICS model to extract domain-invariant and class-specific features as the basis for predictions. We rethink the OOD task in a causal view and highlight both domain-related and class-shared features are confounders that hurt models’ generalization performance. 
To address these challenges, we introduce domain invariance testing (DIT) and class specificity testing (CST) within DICS.
DIT identifies and removes domain-related features from the input, isolating domain-invariant representations. It then evaluates domain invariance by testing the similarity of features within the same class. CST compares current features with multiple learned features, assessing inter-class differences and penalizing instances with high similarity to enhance class specificity.
Extensive experiments on extensive datasets demonstrate the superior performance of DICS on OOD generalization. Additional visualization results further shows that DICS effectively identifies the key features of each class in target domains.

\clearpage
\vfill\pagebreak



\small

\bibliographystyle{IEEEtran}
\bibliography{IEEEabrv,IEEEexample}

\end{document}